\documentclass[letterpaper, 10 pt, conference]{ieeeconf}  

\IEEEoverridecommandlockouts                              

\usepackage{graphicx} 
\usepackage{booktabs}
\usepackage{xcolor}
\usepackage{amsmath} 

\title{\LARGE \bf
Learning Evacuee Models from Robot-Guided Emergency Evacuation Experiments
}

\author{Mollik Nayyar$^{1}$, Ghanghoon Paik$^{2}$, Zhenyuan Yuan$^{3}$, Tongjia Zheng$^{4}$,\\ Minghui Zhu$^{5}$, Hai Lin$^{6}$ and Alan R. Wagner$^{7}$
\thanks{$^{1}$Mollik Nayyar is with the Department of Aerospace Engineering at The Pennsylvania State University
        {\tt\small mxn244@psu.edu}}%
\thanks{$^{2}$Ghanghoon Paik is with the Department of Aerospace Engineering at The Pennsylvania State University
        {\tt\small gip5038@psu.edu}}%
\thanks{$^{3}$Zhenyuan Yuan is with the Department of Electrical Engineering at The Pennsylvania State University
        {\tt\small zqy5086@psu.edu}}%
\thanks{$^{4}$Tongjia Zheng is with the Department of Electrical Engineering at University of Notre Dame
        {\tt\small tzheng1@nd.edu}}%
\thanks{$^{5}$Minghui Zhu is with the Department of Electrical Engineering at The Pennsylvania State University
        {\tt\small muz16@psu.edu}}%
\thanks{$^{6}$Hai Lin is with the Department of Electrical Engineering at University of Notre Dame
        {\tt\small hlin1@nd.edu}}%
\thanks{$^{7}$Alan R. Wagner is with the Department of Aerospace Engineering at The Pennsylvania State University
        {\tt\small alan.r.wagner@psu.edu}}%
}

\begin{document}

\maketitle
\thispagestyle{empty}
\pagestyle{empty}

\begin{abstract}

Recent research has examined the possibility of using robots to guide evacuees to safe exits during emergencies \cite{NayyarWagner2019,NayyarExplanations2020}. Yet, there are many factors that can impact a person's decision to follow a robot. Being able to model how an evacuee follows an emergency robot guide could be crucial for designing robots that effectively guide evacuees during an emergency. This paper presents a method for developing realistic and predictive human evacuee models from physical human evacuation experiments. The paper analyzes the behavior of 14 human subjects during physical robot-guided evacuation. We then use the video data to create evacuee motion models that predict the person's future positions during the emergency. Finally, we validate the resulting models by running a k-fold cross-validation on the data collected during physical human subject experiments. We also present performance results of the model using data from a similar simulated emergency evacuation experiment demonstrating that these models can serve as a tool to predict evacuee behavior in novel evacuation simulations.

\end{abstract}

\section{INTRODUCTION}

Evacuation robots react to alarms by searching for and leading evacuees to exits. Recent research has examined the efficacy of using robots to guide evacuees to safe exits in a building during an emergency ~\cite{robinette2016overtrust,shell2005insights,NayyarWagner2019,NayyarExplanations2020}. The results from these studies demonstrate that evacuees will follow a robot during an emergency. Yet, if robots are going to efficiently guide people to exits during an emergency, then the robot must have some model of how the evacuees will behave. Ideally this model will allow the robot to determine if the evacuee is following it and how best to communicate with the evacuee. Unfortunately, there are few models of evacuee behavior \cite{wang2021incorporating} and no models of evacuee behavior while in the presence of a guidance robot. If we are to create robots that guide evacuees from danger, then models of evacuee behavior that a robot can use to make predictions and adjust its own behavior must be created. This paper presents and evaluates a method for creating evacuee behavior models.

    It is well documented that during an emergency most people calmly move to an exit \cite{gantt2012disaster, vorst2010evacuation}. Few, if any, panic or react irrationally. Moreover, research has shown that during an emergency most people closely comply with directions. We therefore hypothesized that that a model of evacuee behavior could be created which would accurately predict the evacuee's movements during a robot-guided evacuation.
    
    Nevertheless, to be useful the model of evacuee behavior during a robot-guided evacuation must generalize. Ideally, we would like to train an evacuee model on the behavior of a limited number of evacuees and then use that model to predict how all people in the environment evacuate. This paper therefore seeks to contribute a method for learning an accurate, general model of evacuee behavior during a robot-guided evacuation. Moreover, we believe that our method is applicable to other human-robot interaction problems beyond robot-guided emergency evacuation.

    The remainder of this paper begins by discussing the prior related work in this area. Next, we describe the human-subject study that was used to collect data related to the evacuees behavior during a simulated emergency. We then describe the process used to create evacuee models and evaluate the model's accuracy. The paper concludes with a discussion of the results, assumptions, limitations and future work.

\section{Related Work}
There has been substantial work on the mathematical modeling of large-scale evacuations of a populace  \cite{verdiere2014mathematical, provitolo2011emergent, pan2007multi, helbing2000simulating, goatin2009macroscopic, song2014prediction, song2017deepmob, golshani2019evacuation}. Models of how individuals evacuate, on the other hand, are more difficult generate and validate and tend to rely on imperfect and incomplete data or simplistic assumptions \cite{lovreglio2014validation}. For example, the evacuation model developed in \cite{liu2016emergency} assumes that all people follow a prescribed evacuation plan, move at constant speed, and that the exits have unlimited capacity. On the other hand, psychological models of human behavior during emergencies, such as Leach's Dynamic Disaster model \cite{leach1994survival}, tend to be too underspecified to be adapted to simulations of emergency evacuations.   

Social Force models have also be used to simulate evacuee behavior \cite{wei2006evacuation, wan2014research}. Social force models describe pedestrian behavior in terms of acceleration to a desired velocity towards a goal location, a term which maintains a distance from other pedestrians and obstacles, and a term capturing attractive effects \cite{helbing1995social}. These models, however, fail to represent the psychological components that underlie an evacuee's behavior. For example, Leach's model notes that during the initial phases of a disaster, sensory overload and denial will cause approximately 75\% of evacuees to respond to a disaster with inaction \cite{vorst2010evacuation}.  

Moreover, no prior model of evacuee behavior has included the presence of a robot guide. Yet the creation of a model that captures evacuee behavior when being guided by a robot to an exit during an emergency would be a valuable contribution. Such a model could be used to create more accurate robot guided emergency evacuation simulations and contribute to the design of robot evacuation guides themselves.

\section{Emergency Evacuation Data Collection}
The subjects for the experiments were recruited from online study finder advertisement website and using fliers and announcements from a university environment. Some participation requirements included being an adult of at least 18 years of age and capability to climb stairs. Care was take to ensure that the subjects had not been previously recruited for our previous studies. Additionally, close acquaintances of the experimenters were also excluded from the subject pool. For the shepherding experiment we recruited 14 subjects ($50\%$ male; $44\%$ female; $7\%$ non-binary). The average age of the subjects was $28.5$ years old ranging between $20$ and $65$ years old. 


The subjects were pay \$50 for participating and were informed that the experiment involved interacting with a robot but were not told that an emergency would occur. IRB approval was obtained for the experiment. A total of 114 subjects were recruited for the full experiment which included a variety of different experimental and exploratory conditions. The research presented here, however, focuses on only one of the conditions which involved 14 human subject. 

\subsection{Environment Setup}
\begin{figure} [htb]
\centering
\includegraphics[width=3in]{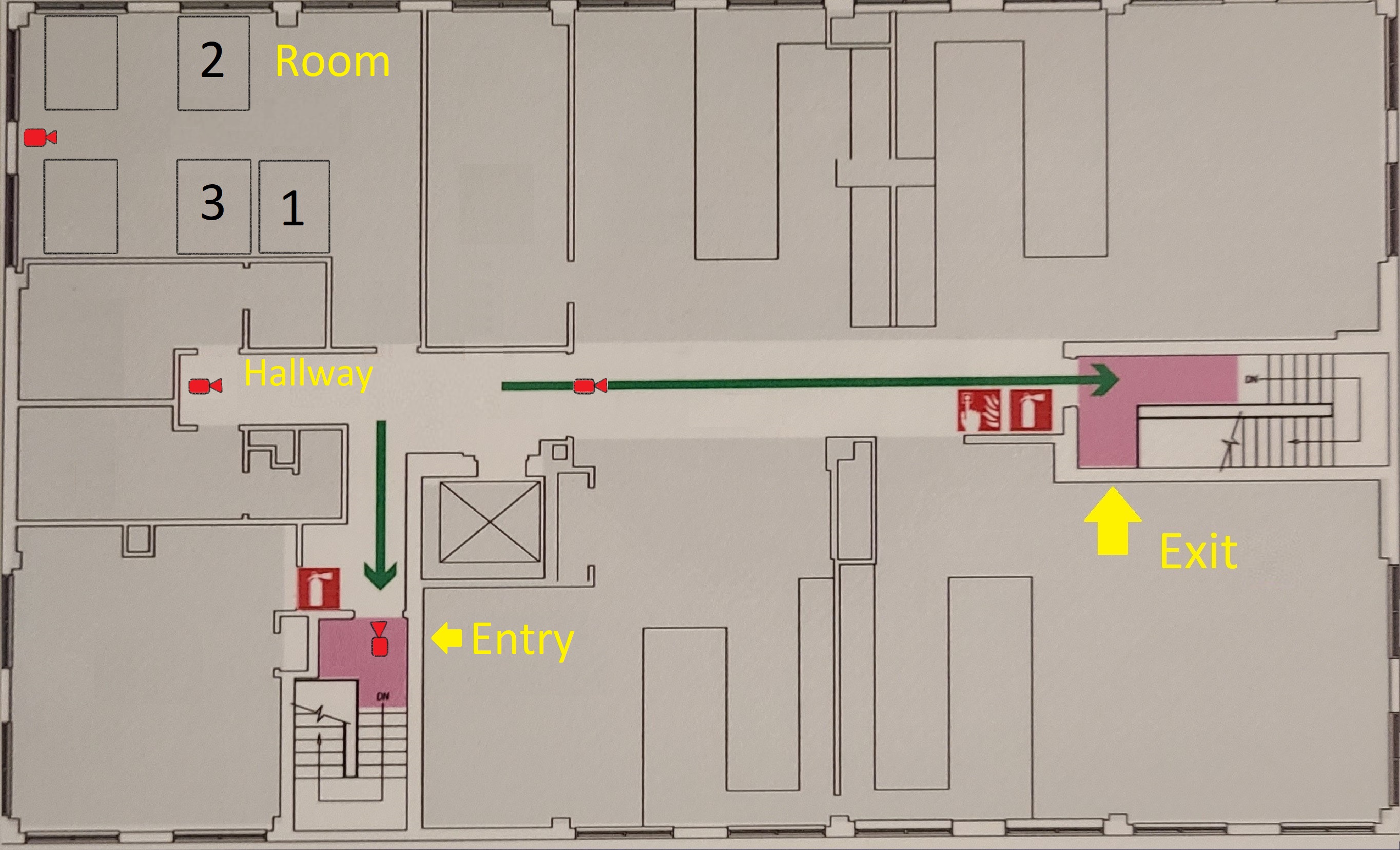}
\caption{The figure shows the layout of the environment used for the experiments. The door marked as `Entry' is the door used to bring the subjects into the environment and was also an exit. As part of the initial guidance, the robot guides the subject to the `Room' to perform the task. The square boxes in the room marked with a 1, 2 or 3 represent the cubicles available for the subject to perform the task. The green arrows depict the two exit pathways available to the subject. After the emergency was initiated, the robot attempts to guide the subject to the point marked as `Exit' following the green arrow. Each of the exits were clearly marked with an exit sign and the subject is not provided any motivation to choose one exit over another. The positions of the cameras are marked with a red camera symbol.}
\label{fig:Layout}
\end{figure}

The experiment was held in an office/lab environment. During the experiment the human subject and the experimenters were the only people present. The floor on which the experiment was performed was outfitted with cameras and Raspberry Pis in order to record the subject’s motion through the environment. Three smoke alarms were also installed that could be controlled by the experimenter. The environment contained the standard exit signs at appropriate locations near the exits and care was taken to ensure no obvious changes to the environment were made. One room was designated as a task room and instructions were posted in this room which directed the subjects to sit at a cubicle and read the paper on the desk of the cubicle. The instructions also reiterated that they would be asked to answer questions about the contents of the paper. A layout of the environment is shown in the Figure \ref{fig:Layout}.

\begin{figure} [htb]
\centering
\includegraphics[angle=-90,origin=c,width=2.8in,height=2.8in]{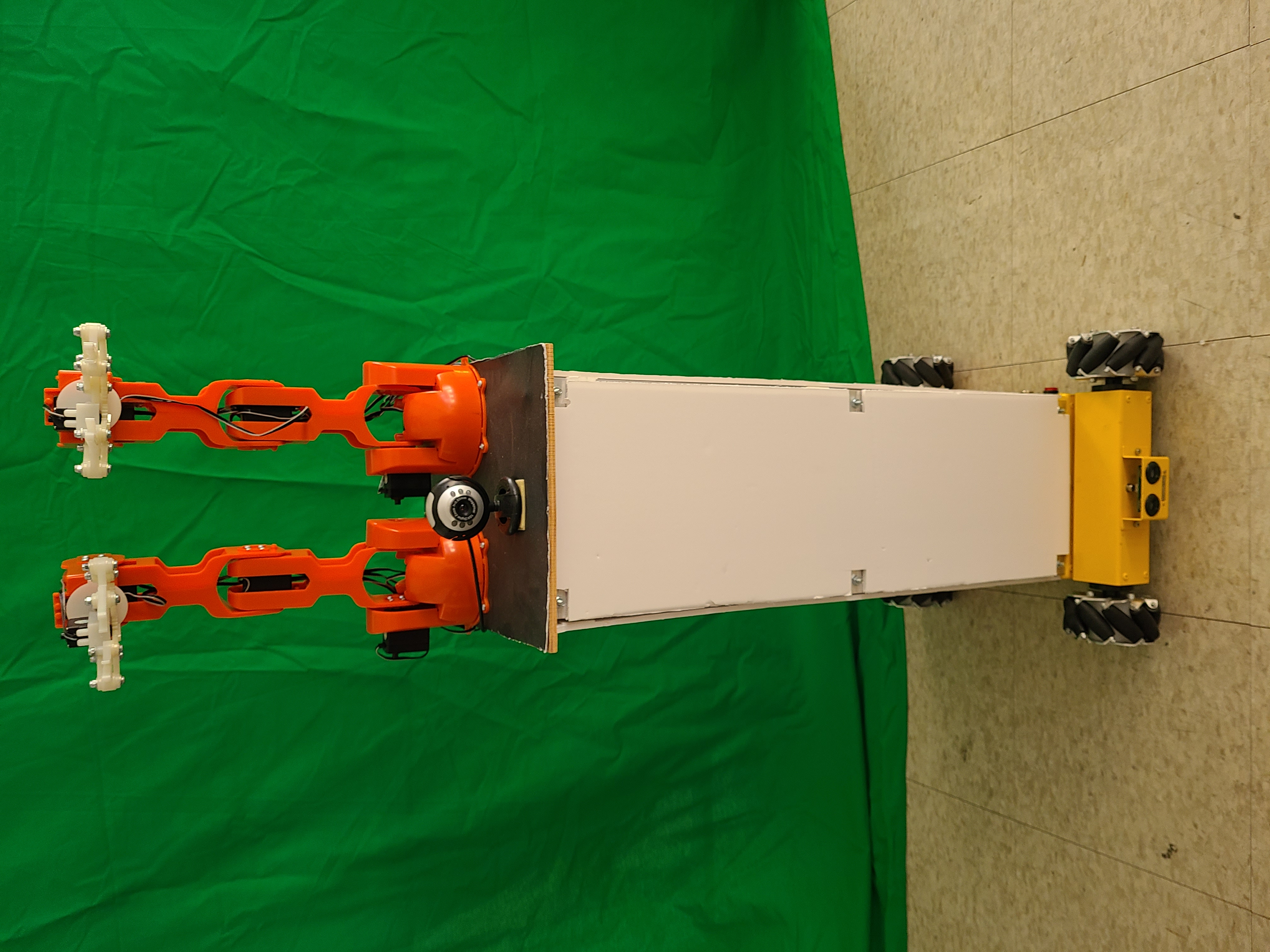}
\caption{The figure shows the robot used for the robot guided emergency evacuation experiment.}
\label{fig:Robot}
\end{figure}

The mobile robot used is depicted in Figure \ref{fig:Robot}. The robot was teleoperated for the duration of the experiment. A 4-wheeled mobile robot was used for the experiment. The robot was outfitted with two arms for the purpose of communicating a sense of direction to the subjects. The robot had a forward facing camera for teleoperation related tasks. The robot was controlled by a Jetson TX2, connected over a local network for communication to the ground station via ROS Kinetic for sending motions and arm gesture commands. The local network was also connected to all of the Raspberry Pis to start and stop the camera recordings. The smoke alarms were manually triggered at the designated time by one of the experimenters.

\subsection{Experimental Procedure}
At the beginning of the experiment, the subjects were greeted by the experimenter at the entrance of the building. Here they were briefed on the task and told that they would be asked to read a paper and answer questions about the paper during the experiment. They were then asked to complete a demographic survey. They were then taken to the designated floor to begin the experiment. At the point marked as `Entry' in Figure \ref{fig:Layout} the experimenter introduced the subject to the robot and told them that the robot would guide them to the task location. The experimenter left once the robot begins to guide the subject to the task location. 

The robot then led the subject to the task room. Upon reaching the room, the robot turned to face the center of the room and continued by using its arms to gesture for the subject to move to the cubicle. Once the subject sat at the cubicle the robot's arms stopped moving but the robot remained at its last location. Next, the subject read the instructions at the cubicle which directed them to read the article in the cubicle. 

After four minutes, which began when the subject sat down, an experimenter triggered the smoke alarms. At the same time, an experimenter also initiates the robot's arms to gesture forward and turns the robot to face the task room door with the arms gesturing towards the door. Ideally at this point the subject begins to follow the robot. The robot then directs the subject towards a more distant exit door which they are unfamiliar with. 

Upon reaching the exit door, the experiment stopped, and the subject was asked to complete a post-experiment survey and was debriefed on the purpose of the experiment. The subject was then compensated for participating and escorted out of the building.

\subsection{Experimental Conditions}
While a number of independent variables were examined as part of this study, since the purpose of this work is to develop an evacuee model for robot-guided evacuation, this paper focuses only on the case of a single robot shepherding a single subject to an exit during an emergency. As detailed in \cite{NayyarWagner2019}, shepherding is a robot guidance method in which a single robot guides evacuees all the way to an exit. Our previous work has shown that shepherding increases evacuee compliance, but because the robot must navigate all the way to an exit, technical challenges arise. In order to mitigate some of the challenges associated with a fully autonomous robot capable of indoor navigation, a teleoperated mobile robot was used for this study. 
Our main objective was to develop a motion model for the evacuees from the captured video data during a robot-guided emergency evacuation.

\section{Evacuee Model Creation}
The data collection setup involved capturing videos of the subject as they moved through the environment during the emergency. In order to use this data, pose estimates were first generated for the human and the robot in the environment. In order to capture the robot's pose, a separate model was trained to detect the robot in the video data.


\subsection{Human Pose Detection}
\begin{figure} [thb]
\centering
\includegraphics[width=2.8in]{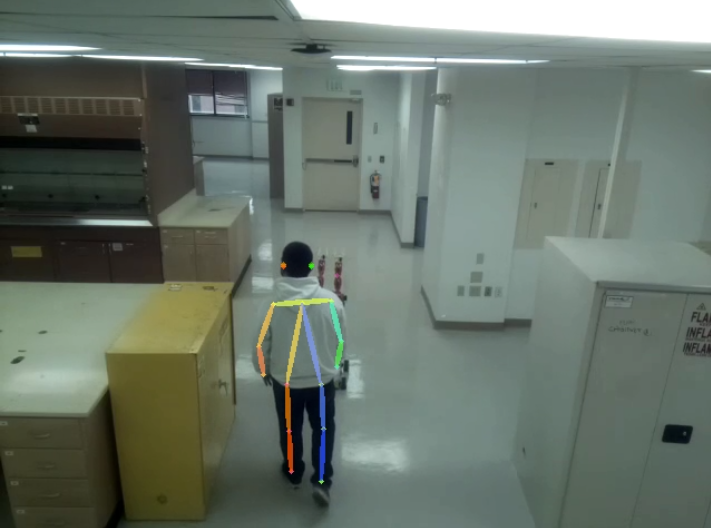}
\caption{The figure shows pose output from the AlphaPose model of a single subject following the robot. The different keypoints representing the body locations can be seen. The ankle keypoints were used for estimating the subject's position in the environment}
\label{fig:PoseDetec}
\end{figure}

Human poses in the environment were extracted using an open source deep learning library called AlphaPose \cite{AlphaPosefang2017rmpe}. The specific AlphaPose model version used in this work is a YoloV3 model \cite{redmon2018yolov3} with a ResNet152 backbone trained on the COCO Keypoint 2017 dataset with 17 body keypoints. The model outputs a JSON file with pixel locations of the detected body keypoints  in each frame along with per keypoint confidence values. This model was used for evacuee pose detection on 640x480 resolution videos collected during the experiment. The left ankle keypoint was used to determine the location of the subject in the environment. Figure \ref{fig:PoseDetec} shows an example of the pose detections of one of the subjects during the experiment.

\subsection{Camera – World Space conversion}
\begin{figure} [tb]
\centering
\includegraphics[width=2.8in]{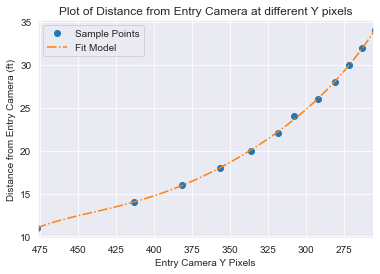}
\caption{The figure shows model used to convert camera pixels to a distance measurement in feet. This is a representative plot for the Entry camera. The X-axis of the plot represents the pixels in the image frame along the y (or vertical) direction of the image frame. The Y-axis of the plot is the distance in feet from the camera.}
\label{fig:camModel}
\end{figure}

\begin{figure} [h]
\centering
\includegraphics[width=2.8in]{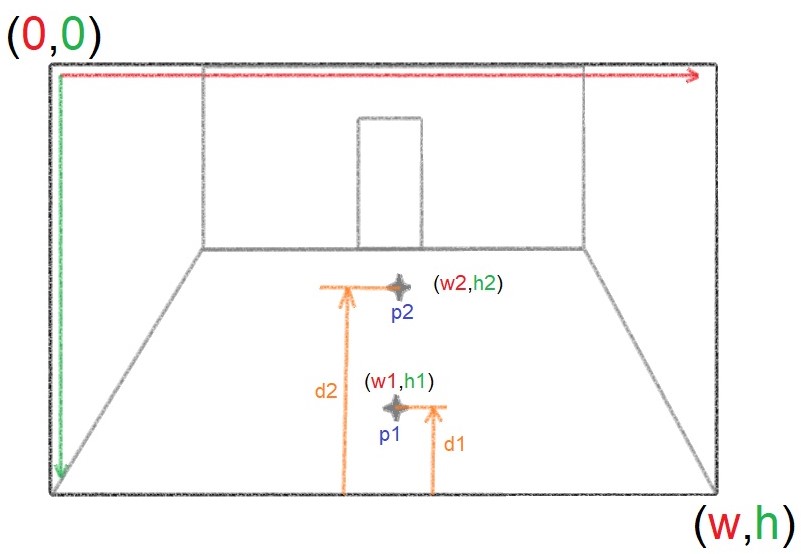}
\caption{The figure shows a representative scene depicted on a notional camera frame. The conventional coordinates of the camera are shown in red and green for width and height respectively. For some arbitrary pixels on the frame (p1 and p2), the ground truth distance from the camera was measured and stored (d1 and d2). These distances at specific pixels were then used to create a polynomial model. This model could then convert any arbitrary pixel to a distance value from the camera. A similar approach was taken for the horizontal direction of the camera as well but instead of absolute distances, the width in pixels and the corresponding width in units of measurement were used to fit the model.}
\label{fig:camFrame}
\end{figure}

\begin{figure}[b]
\centering
\includegraphics[width=2.8in]{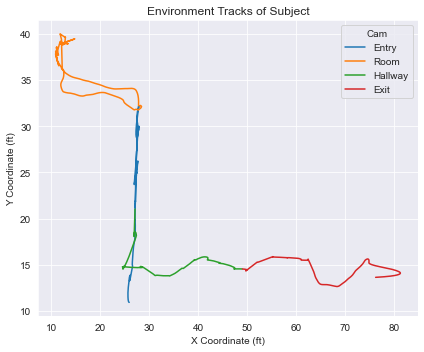}
\caption{The figure shows a global track of a single subject in the shepherding condition. The different colors correspond to the tracks generated from the different cameras placed in the environment. The blue color is the `Entry' camera, the orange color is `Room' camera, the green track is from the `Hallway' camera and the red track is from the `Exit' camera}
\label{fig:EnvironmentTrack}
\end{figure}

In order to develop a generalizable evacuation system capable of being used in a variety of different environments, evacuee videos were collected from notional surveillance cameras placed in the hallways.
The videos collected from the camera system were used to obtain the world space locations of the subjects. To convert the pixels of the detected keypoints to the world space, calibration images were collected in the environment at known ground truth distances from the cameras and the AlphaPose model was used to obtain the keypoint locations in pixels. These distance-pixel calibrations were used to create a polynomial model relating the image frame y-axis pixel location to the distance from the camera as shown in Figure \ref{fig:camFrame}. For the X-axis (or horizontal distance in the image frame), the width of a calibration object in pixels and in inches taken at different y-axis was used to create a model for the X direction. Combining the two, we obtained the coordinates of the subject with respect to the camera. This was then transformed into world space coordinates by incorporating the camera’s world space coordinates. This conversion was performed for each of the cameras to obtain a global track of the subject in the environment as shown in Figure \ref{fig:EnvironmentTrack}. An example of the camera model is shown in Figure \ref{fig:camModel}. For the `Entry' camera as shown in Figure \ref{fig:Layout}, the positive $X$ and $Y$ axes of the camera aligns with the positive $X$ axis and negative $Y$ axis in the world space, however, this relation is reversed for the `Room', `Hallway' and `Exit' camera, i.e. the positive $X$ of the camera corresponds to the negative $Y$ axis of the world frame and the positive $Y$ axis of the camera corresponds to the positive $X$ axis of the world frame. This change in orientation of the cameras was accounted for before making the relevant coordinate transformations for the subject positions.


\subsection{Robot Detection and Tracking}
To obtain the robot’s world space positions, a YoloV5 model \cite{Yolov5glenn_jocher_2022_7002879} was trained with labeled images of the robot in the videos with its bounding boxes. The bounding box detections were converted to a single pixel point in the frame representing the location of the robot in the camera frame. The camera to world space conversion models were used to obtain the positions of the robot. In many cases, the robot was occluded by the subject in the experiment in which case the location was estimated as a linear interpolation of the pixels between the two successful detections.

\subsection{Savitzky–Golay filter}
The tracks obtained from the camera models tended to be noisy. A small change in the detected pixel from the AlphaPose model resulted in a large change in the estimated position. To mitigate potential issues in evacuee motion model accuracy due to this noise, data filtering was performed to clean out the noise and jitter. A locally weighted polynomial regression method called Savitzky–Golay filter was used for this task \cite{savitzky1964smoothing}. The filter is commonly used for digital signal smoothing and derivative calculations. The filtered plot of the subject motion in the shepherding condition is shown in Figure \ref{fig:EnvironmentTrack}

\subsection{Time Series Modeling}
\label{TimeSeriesModeling}
The data obtained after the conversion and filtering process contained positions of the subject and the robot, from each of the cameras, in units of feet. The positions were available at each $0.025$ seconds or at a frequency of $40$ frames per second. An evacuee model should be able to predict the next position of the subject as the output dependent on some choice of input variables. This problem could be conveniently designed as a time series prediction problem where the \textit{future} $X-Y$ positions of the subject is treated as the output and the \textit{past} positions of the robot, the subject and their relative distance are the inputs. It was decided to model this as a supervised learning problem with augmented autoregressive features of the positions at a desired time lag. The choice of time lags and the number of lags then become hyperparameters for tuning the model. Modeling the problem as a supervised learning problem allows us to utilize a large variety of regression algorithms.
\begin{table}[!ht] 
\centering
    \begin{tabular}{cccc} 
    \toprule
     $X_{s}{l10}$ &  $Y_{s}l10$ &  $X_{r}l10$ &  $Y_{r}l10$ \\
    \midrule
  0.210581 &   0.701602 &   0.115875 &   0.897624 \\
  0.018134 &   0.972653 &   0.055834 &   0.921662 \\
  0.016457 &   0.953926 &   0.086825 &   0.921233 \\
     \vdots & \vdots & \vdots & \vdots\\
  0.022875 &   0.983208 &   0.117549 &   0.897630 \\
  0.003972 &   0.932113 &   0.116921 &   0.896623 \\
  0.012397 &   0.917920 &   0.086825 &   0.921233 \\
    \bottomrule
    \end{tabular}
    \caption{Example of input data format}
    \label{Table:inputData}
\end{table} 

To convert the data for use in a supervised learning paradigm, a dataset was created containing the $X-Y$ positions of both the subject and the robot at the current time $t_K$ for each subject. This data was then augmented with the past positions of the subject and the robot and their relative distance at time $t_{K-m}$ where $m$ is the desired lag hyperparameter. Additionally, in order to remove the effect of the scale of distances from the data, the features were normalized using a Min-Max scaler to bring the values between 0 and 1. Data scaling was also needed to perform validation on simulation experiments as discussed in section \ref{sec:simEval}. Table \ref{Table:inputData} shows a sample input data frame. The columns labeled $X_{s}{l10}$ and  $Y_{s}l10$ refers to the subject's $X$ and $Y$ position with a lag of 10 frames respectively. Since the videos of the subjects were captured at 40 frames/second, a lag of 10 frames corresponds to a lag of 0.25 seconds. Therefore, we provide the model of the subject and robot's positions 0.25 seconds prior to the predicted position of the subject. 


For the purposes of solving the time series prediction problem using a supervised learning approach, we use the XGBoost algorithm \cite{chen2016xgboost}. 
XGBoost is a gradient boosting algorithm that has achieved state-of-the-art performance on a wide range of tasks, including regression, classification, and ranking \cite{bentejac2021comparative}. The algorithm works by iteratively adding new weak learners to the model, with each new learner attempting to correct the errors of the previous learners \cite{chen2016xgboost}. The model was trained on the data collected from the experiment with augmented autoregressive features of the positions of the robot and the subject with a lag of $10$ frames or $0.25$ seconds. 


\section{Model Evaluation}
The procedure described in the previous section was run on 14 unique human subjects. We will refer to the data from each of these experimental runs as set 1 thru 14. Out of the 14 runs, data from 2 runs could not be used as the subject did not follow the robot. For convenience, each of the the subjects were numbered from 1 to 14 and the data from their corresponding experiments was named as 'Exp'+`number of the subject'. This results in our experimental dataset named as 'Exp1' for subject 1 and so on for each individual subject. It needs to be noted that subject 1 and 6 did not follow the robot and hence were not included in the training or testing data. To evaluate the performance of the algorithm, we perform a k-fold validation procedure with $k$ being the number available cases. 
This resulted in a total of 12 different models and the results from the evaluations were aggregated. The trained regression model obtained from the XGBoost algorithm was evaluated by comparing the performance of the model on a similar but unseen data from the same environment with the same underlying parameters of scale, speed and exit paths. We also took the trained model and test it on data collected from an online emergency evacuation study that was conducted in a simulated Unity environment two years ago. The online study features the same experimental setup of emergency robot-guided evacuation but completely different experimental parameters with it being a simulation in different environment, at a different scale, and with  different robot and human motion speeds. 


\subsection{k-fold Validation}
\begin{figure}[htb]
\centering
\includegraphics[width=3in]{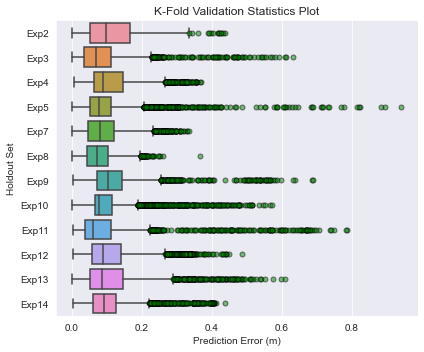}
\caption{The figure shows box plots for the prediction error of the evacuee position from the k-fold validation for each of the holdout sets. The error values are in units of meters}
\label{fig:Dist_errors}
\end{figure}

The results of the models on the holdout set are presented in the Table \ref{Table:holdoutResults}. The results were converted to meter scale. The data from the holdout sets were not used for training. In other words, the result in $1^{st}$ row represents the prediction error of the trained model on the data from the $2^{nd}$ run of the experimental condition whereas the trained model is created from the data from \textit{all other} cases. The Figure \ref{fig:Dist_errors} shows the error statistics for all the holdout sets as a box plot. It is evident that while there are outliers with large errors, the majority of the samples predicted are very close to the small range around the mean. The errors statistics are presented in the Table \ref{Table:holdoutResults}. $\mu_{e}$ represent the \textit{mean} of the L2 norm between the predicted position and the position from the video data. $\sigma_{e}$ represent the standard deviation of the errors from the mean. These results show that the mean error in position is small, with the mean error measuring approximately 9.9 centimeters.


\begin{table}[htb]
    \centering
\begin{tabular}{cll}
\toprule
 Holdout Set & $\mu_{e}$ (m) & $\sigma_{e}$ (m)\\
 \midrule
2  &     0.109 &   0.067 \\
3  &     0.087 &   0.074 \\
4  &     0.112 &   0.070 \\
5  &     0.095 &   0.081 \\
7  &     0.090 &   0.057 \\
8  &     0.077 &   0.044 \\
9  &     0.117 &   0.079 \\
10 &     0.098 &   0.064 \\
11 &     0.090 &   0.095 \\
12 &     0.109 &   0.076 \\
13 &     0.108 &   0.083 \\
14 &     0.103 &   0.066 \\ \hline
mean  &    0.099 &   0.071 \\
\bottomrule
\end{tabular}
\vspace{2mm}
    \caption{Resulting error in position prediction for the holdout sets}
    \label{Table:holdoutResults}
\end{table}

\begin{figure}[h]
\centering
\includegraphics[width=2.8in]{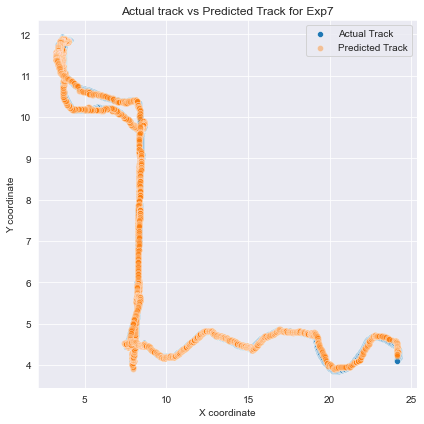}
\caption{The figure shows the actual and the predicted track of the subject in dataset 7. The close correspondence of the predicted track with the observed track demonstrates the ability of the model to provide an accurate prediction of the subject's position at a future time step. The distance on the axes are in meters.}
\label{fig:XYerrors}
\end{figure}


\subsection{Evaluation on Simulation Data} \label{sec:simEval}
\begin{figure}[htb]
\centering
\begin{minipage}{.38\textwidth}
\includegraphics[width=\linewidth]{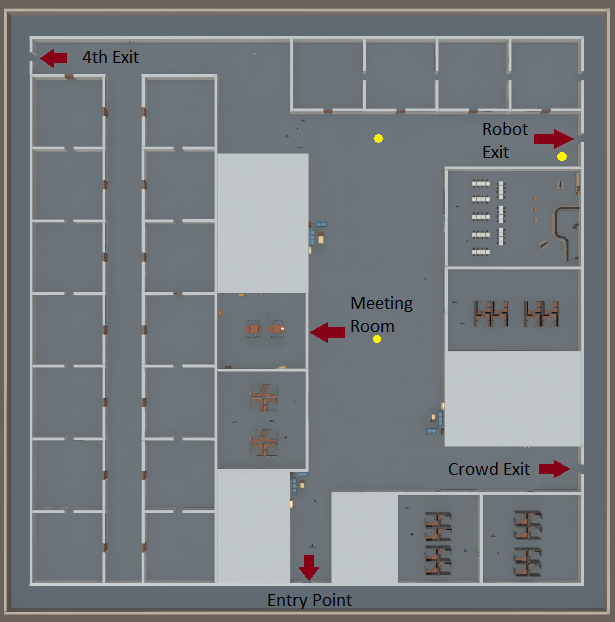}
\caption{The figure shows the layout of the simulated environment. The coordinates of the tracks from the simulation needed to be changed to be in the same frame as the data collected from the physical experiments. The simulation experiment follows the same general structure as the physical human evacuation experiments. The distance in the simulation are in units of meters}
\label{fig:simEnv}
\end{minipage}%
\vspace{3mm}
\begin{minipage}{.38\textwidth}
\includegraphics[width=\linewidth]{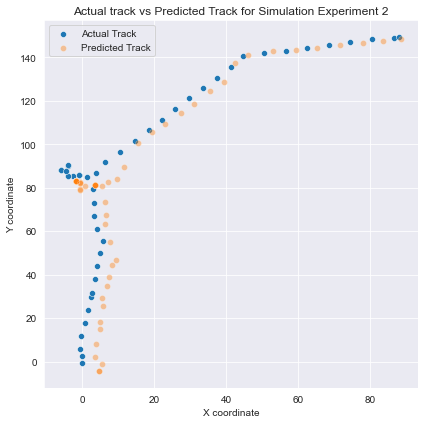}
\caption{The figure shows the tracks of the subject and prediction model in the simulated environment. The one step future prediction models give good correspondence with observed tracks. The distance in the simulation are in units of meters}
\label{fig:sim2Track}
\end{minipage}
\end{figure}

In order to examine the applicability of the models in other environments, we decided to use the models to predict human subject motion during a similar task performed in a 2D simulated environment. The environment was created using the Unity game engine. A robot guided evacuation scenario was again used. The simulated experiment featured a much larger environment designed with different units of measurements, different robot speeds and different directions towards the exits. The layout of the environment can be seen in the Figure \ref{fig:simEnv}. The data collected was at a much slower frequency of 1Hz. No part of the simulation dataset was used in training the models. In order to ensure proper correspondence with the training data, the positions of the subject and robot were transformed into the proper coordinate frame. It should also be noted that the simulation environment used different units of measurement for distances and positions. This discrepancy was resolved by the data normalization process described in the Section \ref{TimeSeriesModeling}. The simulation data was augmented with the lags of the subject and robot's positions but due to the limited data available from the simulations, the smallest lag that could be achieved was one second. It was observed that changing the lag during the training process to 20 frames or even 40 frames yielded poor performance on the simulation data. We therefore trained the motion model with a lag of 10 frames. 

\begin{table}[thb]

    \centering
\begin{tabular}{lrr}
\toprule
Experiment &  $\mu_{e}$ (m) & $\sigma_{e}$ (m)\\
\midrule
Sim. Exp 1 &     6.322 &   1.654 \\
Sim. Exp 2 &     8.919 &   2.173 \\
Sim. Exp 3 &     3.688 &   2.834 \\
Sim. Exp 4 &     7.109 &   2.574 \\
Sim. Exp 5 &     7.480 &   2.015 \\ \hline
mean  &     6.703 &  2.250  \\

\bottomrule
\end{tabular}
\vspace{2mm}
    \caption{Resulting error in position prediction for the simulation sets.}
    \label{tab:resultsSim}

\end{table}

The results are presented in Table \ref{tab:resultsSim}. The table shows that the mean absolute errors for the predicted subject's positions in the simulation when tested on the model generated from the physical evacuation experiment. The mean errors were within $7$ meters with a standard deviation of $2.25$ meters. The higher error can be attributed to noisy camera model calibration in for the horizontal axis of the camera. Considering the vastly different nature of data compared to the training set, it can be argued that the predictions are reasonable. 

A full track depicting the prediction results from the model can be seen in Figure \ref{fig:sim2Track}. Notice that the track of the one-step prediction model does reasonably well on the simulation data. It should be noted that the simulation data contained fewer samples compared to the training dataset from the physical experiments. Although the model performs worse on the simulation data, it nevertheless does a generates a reasonable prediction of the subject's track given the difference in the scale of the environment, the units of measurements and the speed of the robot.



\section{Conclusions}
    The paper describes and evaluates a process for creating predictive behavioral models of human subjects during a robot-guided emergency evacuation. The process uses camera data collected in a ecologically valid manner from a notional camera surveillance system. Our results show that models created from the human evacuation data can be used to create a prediction model of how evacuees will follow the robot with average positional error of less than 10 centimeters. A much more challenging out of distribution evaluation results in error of approximately 6.7 meters with tracks that closely correspond to the subject's movements. 
    
    This paper makes several valuable contributions. First, we demonstrate a method for collecting and using human subject behavioral data during robot-guided emergency evacuations to create highly accurate predictive models of how evacuees will behave and move during an emergency. Second, we show that these evacuee models accurately generalize across people within the same environment. This is a shocking result. One might naively predict that evacuation is a chaotic process and how an evacuee will follow the guidance of a robot makes evacuee behavior unpredictable. Yet our experiments, data, and models show otherwise. Finally, we demonstrate that, although some predictive accuracy is lost, these models can make predictions across environments. We believe that our approach can be used model other human-robot interaction situations and to create models that can then be used to inform the robot how people will behave in other situations. 
\section{Limitations}
    Our approach does have limitations. Perhaps most importantly, although we have shown that our model generalizes to a new environment, it is unclear how different that environment can be before the model accuracy drastically diminishes. Although our approach makes assumptions, we have intentionally tried to minimize the impact of these assumptions. For example, using realistically placed cameras to capture the data. Being able to create accurate evacuee models bring us a step closer to our ultimate goal of creating robots capable of guiding people to safety.

\section*{ACKNOWLEDGMENT}

This material is based upon work supported by the National Science Foundation under Grant Number CNS-1830390 and IIS-2045146. Any opinions, findings, and conclusions or recommendations expressed in this material are those of the authors and do not necessarily reflect the views of the National Science Foundation.


\bibliographystyle{IEEEtran}
\bibliography{root}

\end{document}